\pdfoutput=1

\documentclass[11pt]{article}

\usepackage{naacl2021}

\usepackage{times}
\usepackage{latexsym}
\usepackage{algorithmic, algorithm}

\usepackage[T1]{fontenc}

\usepackage[utf8]{inputenc}

\usepackage{microtype}

\usepackage{graphicx} 

\usepackage[symbol]{footmisc}

\usepackage{enumitem}
\setlist{nosep, leftmargin=5mm} 

\usepackage{letltxmacro}
\LetLtxMacro\oldttfamily\ttfamily
\DeclareRobustCommand{\ttfamily}{\oldttfamily\csname ttsize\endcsname}
\newcommand{\setttsize}[1]{\def\ttsize{#1}}%

\title{Robustly Optimized and Distilled Training for Natural Language Understanding}

\author{Haytham ElFadeel \\ \texttt{haythamf@fb.com} \\\And Stan Peshterliev \\ \texttt{stanvp@fb.com}
    \\}

\begin{document}
\maketitle
\begin{abstract}
In this paper, we explore multi-task learning (MTL) as a second pretraining step to learn enhanced universal language representation for transformer language models. We use the MTL enhanced representation across several natural language understanding tasks to improve performance and generalization. Moreover, we incorporate knowledge distillation (KD) in MTL to further boost performance and devise a KD variant that learns effectively from multiple teachers. By combining MTL and KD, we propose Robustly Optimized and Distilled (ROaD) modeling framework. We use ROaD together with the ELECTRA model to obtain state-of-the-art results for machine reading comprehension and natural language inference. 
\end{abstract}

\setttsize{\small}%

\section{Introduction}

Machine reading comprehension (MRC) aims to answer questions from unstructured text~\citep{hermann2015teaching, joshi2017triviaqa, rajpurkar2018know}, which is a longstanding goal of natural language understanding (NLU). Another important NLU task is natural language inference (NLI) which recognizes if a hypothesis sentence entails, contradicts or is neutral with respect to a premise sentence. Since the introduction of the BERT language model with self-supervised pretraining~\citep{devlin2018bert}, there have been many new state-of-the-art results in MRC and NLI, and more broadly a resounding impact on the way contemporary language modeling is carried out.

BERT demonstrated that leveraging large amounts of unlabeled data is effective for learning universal language representation. Another approach to learn universal language representation is multi-task learning (MTL) where a single model is trained on multiple supervised tasks at once, such as MRC, NLI, and sentiment analysis. MTL is inspired by human learning where people learn to use language by performing multiple tasks (e.g. reasoning, making statements, asking and answering questions), and often knowledge learned from one task is used to learn other tasks. 

In addition to pretraining, a recent successful method for improving performance is knowledge distillation (KD). KD transfers knowledge from a ``teacher'' model to a ``student'' model by training the student to imitate the teacher’s outputs. Excellent results have been archived with KD knowledge transfer to the same-size models, and smaller-size models for computational efficiency. However, transferring knowledge from multiple teachers to a student is less studied.

In this paper, we investigate MTL and KD as a second pretraining step to improve the universal language representation, downstream performance, and generalization in NLU tasks. Moreover, we propose a novel variation of KD that can combine information from multiple teachers and even exceeds the performance of the multiple teacher ensemble. Finally, we incorporate MTL and KD approaches in Robustly Optimized and Distilled (ROaD) modeling framework. ROaD training together with the ELECTRA~\citep{clark2020electra} model obtains new state-of-the-art results in MRC and NLI tasks.

Our contributions are as follows: 
\begin{itemize}
\item We propose and evaluate a new MTL pretraining approach that yields performance and generalization improvements to self-training alone and prior work on MTL pretraining.
\item To improve KD from multiple teachers, we devise a novel teachers weighting schema and demonstrate that it yields improvements compared to any individual teacher and even the ensemble of teachers.
\item We present ROaD-ELECTRA models for MRC and NLI models that yield significant improvements and achieve new state-of-the-art results on SQUAD 2.0 and MNLI.
\end{itemize}

\section{ROaD Modeling Framework}
\subsection{Multi-task Learning Pretraining}
Transformer language models are pretrained in a self-supervised manner to predict a masked token in a context, MLM objective, and predict the next sentence, NS objective. Recently, the ELECTRA model added a new objective to distinguish "real" input tokens vs "fake" input tokens generated by another neural network. Self-supervised pretraining objectives allow the models to learn about language structure and semantic from large datasets, but they do not take advantage of the available supervised data that exists for multiple NLU tasks.

MTL pretraining aims to improve upon self-supervised pretraining using multiple supervised objectives to build language representation~\citep{zhang2017survey, liu2019multi}. The goal of ROaD MTL pretraining is to teach the model a diverse set of NLU tasks to help it build robust universal language representation and generalize better. By training on multiple tasks, we reduce overfitting since the model representation has to generalize across diverse tasks.

\textbf{Training Procedure.} For ROaD MTL pretraining, we start from a transformer model that is pretrained with self-supervision such as BERT and ELECTRA. Then, we perform the MTL pretraining on diverse NLU tasks and datasets.


We have a separate prediction \textit{head} for every task with shared universal representation. We use the Adam optimizer to update the parameters of all task-specific heads and the shared representation layers. In every training step, we perform multiple passes (forward, compute loss and gradient accumulate) using a randomly selected datasets. At the end of the step, we average the gradient across tasks and make a parameter update. Details are in Algorithm~\ref{mt_pretraining_alg}.

\begin{algorithm}[H]
\small
\caption{ROaD MTL pretraining procedure}

\begin{algorithmic}
\STATE step-max : max number of steps 
\STATE pass-step : number of datasets to use per step (also the gradient accumulation)
\\\hrulefill
\FOR {step in 1, 2, 3, step-max}
\FOR {pass in 1, 2, pass-step}
\STATE dataset $\gets$ random(datasets)
\STATE B$_{ds}$ $\gets$ get\_minibatch(dataset)
\STATE Compute Loss
\STATE Compute and Accumulate Gradient
\ENDFOR
\STATE Average Gradient
\STATE Update model
\ENDFOR

\end{algorithmic}
\label{mt_pretraining_alg}
\end{algorithm}

\textbf{Tasks and datasets.} For ROaD multi-task pretraining, we selected nine diverse publicly available datasets and categorized them into four tasks, each with a separate prediction head, see Table~\ref{mtl_dataset}. The tasks are MRC, NLI, sentiment analysis (SA), and paraphrase identification (PI).

\begin{table}[!ht]
\small
\centering
\setlength{\tabcolsep}{0.5em} 
{\renewcommand{\arraystretch}{1.3}
\begin{tabular}{p{0.24\linewidth}p{0.08\linewidth}p{0.45\linewidth}p{0.08\linewidth}}
\hline
\textbf{Dataset} & \textbf{Task} & \textbf{Description} & \textbf{Size} \\ \hline
TriviaQA~\citeyearpar{joshi2017triviaqa} & MRC & Questions and answers from trivia enthusiasts & 114K \\
Natural Questions (NQ)~\citeyearpar{kwiatkowski2019natural} & MRC & Questions from search logs with answers from Wikipeida. We use the short answers & 204K \\
QuAC~\citeyearpar{choi2018quac} & MRC\footnote{with additional heads for followup and yes/no} & QA with context in a dialog. & 83K \\
QA-SRL~\citeyearpar{fitzgerald2018large} & MRC & QA driven semantic role labeling & 172K \\
QA-ZRE~\citeyearpar{levy2017zero} & MRC & QA driven relation extraction & 1000K \\
MNLI~\citeyearpar{mnli} & NLI & Multi-Genre NLI & 392K \\
SNLI~\citeyearpar{snli:emnlp2015} & NLI & Stanford NLI & 549K \\
Amazon~\citeyearpar{ni2019justifying} & SA & Sentiment analysis on reviews. Predict the rating. & 1000K \\
QQP\footnote{https://www.quora.com/q/quoradata/First-Quora-Dataset-Release-Question-Pairs} & PI & Quora duplicate question paraphrase identification & 363K \\ \hline
\end{tabular}
}
\caption{\label{mtl_dataset}
ROaD MTL pretraining datasets and tasks.
}
\vspace{-3mm} 
\end{table}

\textbf{Relationship to MT-DNN and T5.} MT-DNN~\citep{liu2019multi} combines BERT with second MTL pretraining on supervised tasks to achieve improved performance on several NLU tasks. ROaD approach to MTL pretraining is similar to MT-DNN but it differs in three ways. (1) We choose datasets and tasks geared towards MRC and NLI. (2) We average the gradient from a few randomly selected tasks for each step. While MT-DNN performs a step using the gradient of a single task. (3) We use the same prediction head for each task across datasets whereas MT-DNN uses a separate head per dataset. For example, the MRC tasks such as TriviaQA, NQ and QuAC share the same prediction head.

T5~\citep{raffel2019exploring} is a seq2seq model that uses self-training and multi-task supervised training at the same time.  The encoder-decoder architecture gives T5 flexibility to naturally model diverse problems like machine translation and summarization. In our work, we focus on encoder-only architectures, like BERT and ELECTRA, since they are strong at our target MRC and NLI tasks while being computationally efficient for production deployment.

\subsection{Knowledge Distillation Pretraining}


\begin{figure}[!ht]
\centering
\includegraphics[width=\columnwidth]{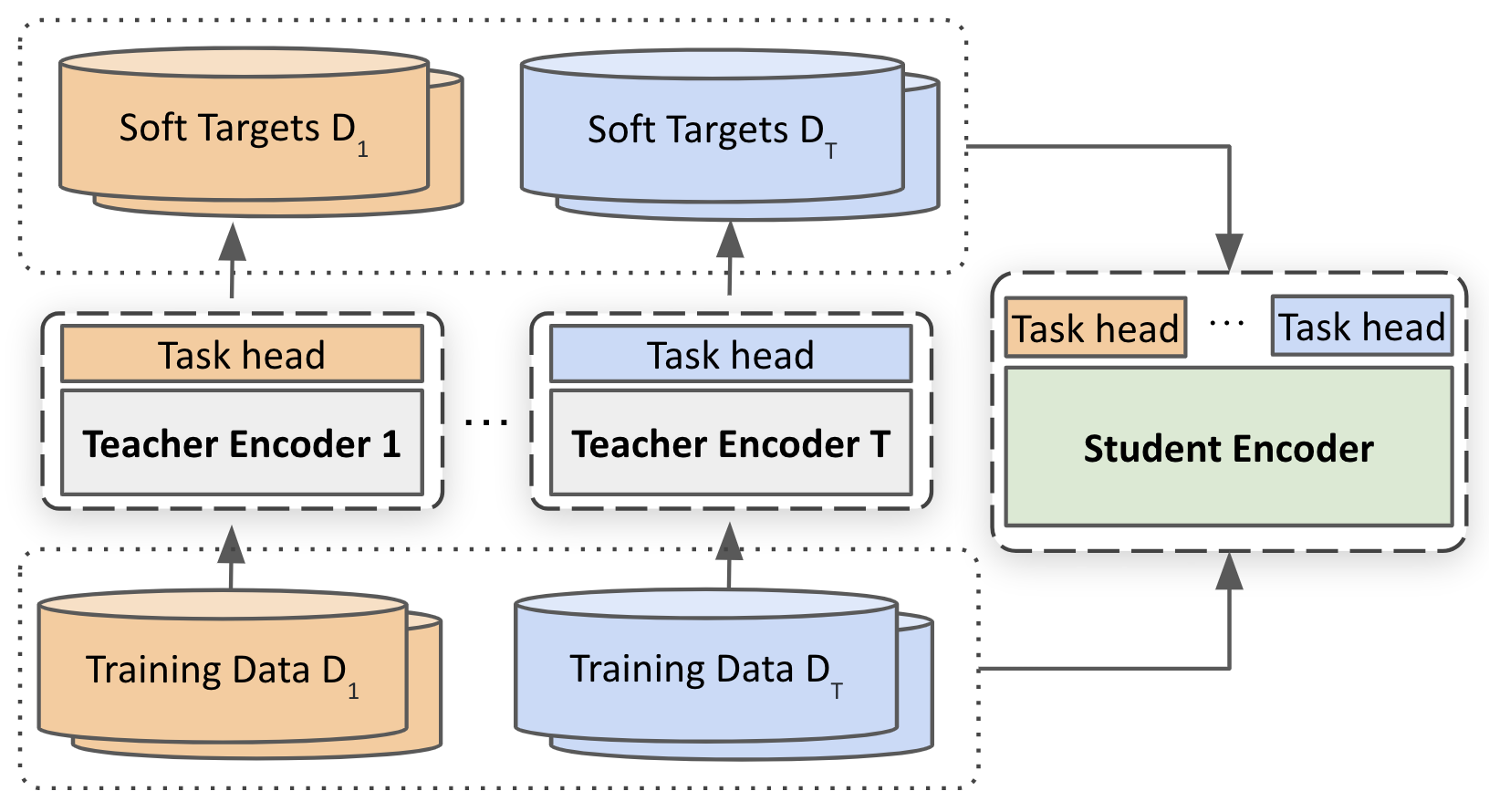}
\caption{
ROaD KD pretraining procedure. Fist, finetune teachers on each training dataset $D_T$. Then, use the teacher to generate soft targets for each dataset. Finally, use the generated soft targets with existing hard targets to pretrain a ROaD representation student model.
}
\label{fig:road_kd}
\end{figure}


KD is done by using the teacher output logits or feature information. We use logits~\citep{hinton2015distilling} with a loss function that is a weighted average of two objectives. The first objective is the original cross entropy (CE) with ground truth ``hard targets''. The second objective is the Kullback–Leibler (KL) divergence with ``soft targets'' targets from the teacher predictions. For the second KL objective, we use softmax with high temperature $T > 3$ to generate a softer probability distribution over classes. We set the same temperature in the teacher and student model. See Figure~\ref{fig:road_kd}.

\textbf{KD and MTL}. To improve MTL pretraining we introduce KD as an additional objective. Figure 1, First, for each dataset we finetune a model as a teacher using the dataset training data. Then, we generate a set of teacher soft targets for each dataset. During the MTL pretraining, we define the objective $J$ to be a combination of CE and KD where $\lambda$ is a weight for the strength of the KD term. %
\vspace{-5px}%
$$J = (1 - \lambda) CE(\textit{hard targets}) + \lambda KL(\textit{soft targets})$$

\subsection{Knowledge Distillation in Finetuning}

\begin{figure*}[!ht]
  \centering
  \begin{minipage}[c]{0.67\textwidth}
    \includegraphics[scale=0.31]{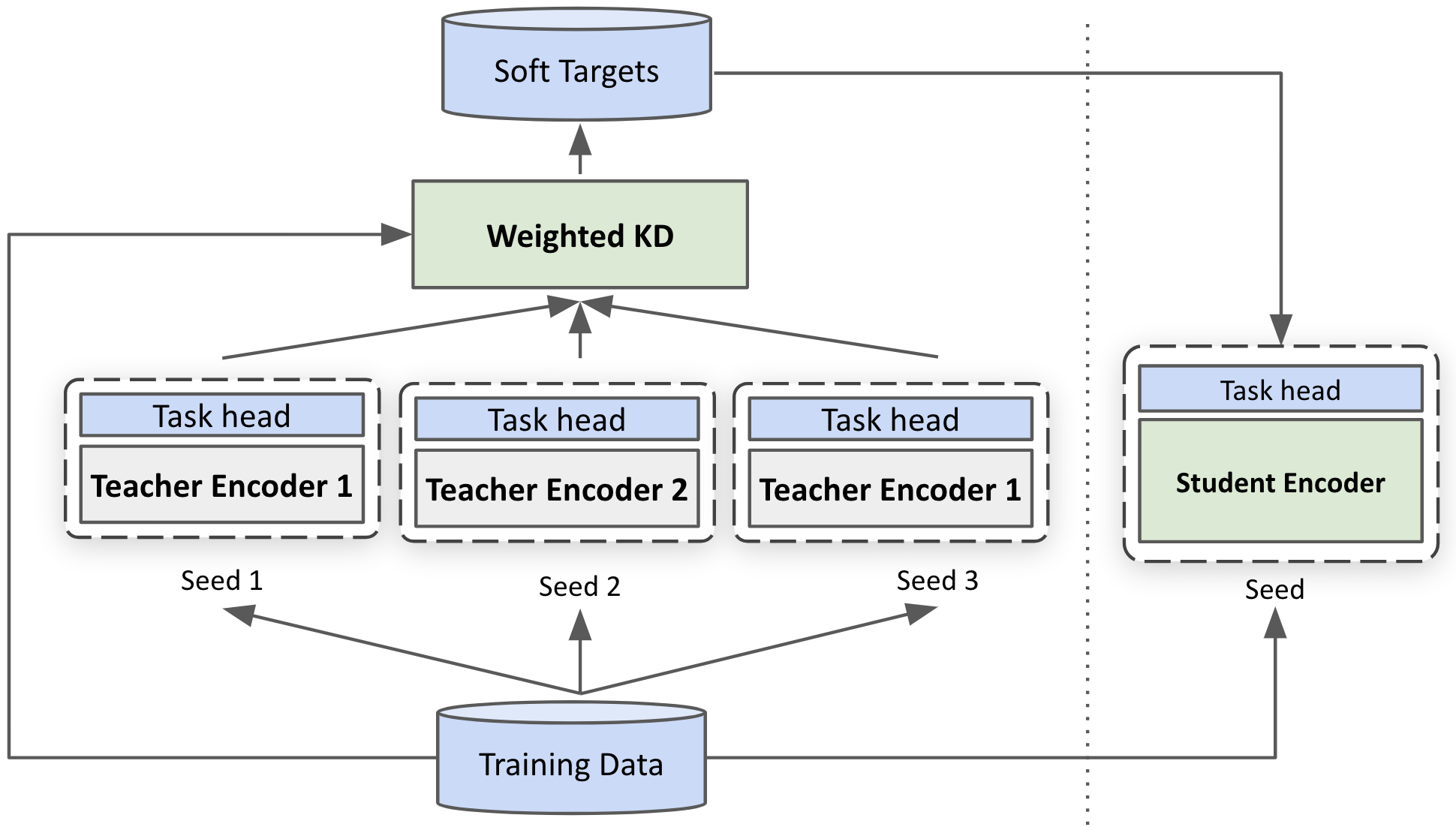}
  \end{minipage}\hfill
  \begin{minipage}[c]{0.3\textwidth}
    \caption{
ROaD Weighted KD finetuning procedure. First, finetune $N$ ROaD MTL models as teacher ensemble on a task using different seeds. Then, collect the logits of the $N$ models combined with hard targets, and use Weighted KD to generate soft targets. Finally, use the generated soft targets with hard targets to train the final ROaD model.
} \label{fig:weighted_kd}
  \end{minipage}
\end{figure*}

During model finetuning, we can improve KD by using an ensemble of multiple teachers. However, the student has to take advantage of the strengths of the different teachers. Thus, we propose \textit{Weighted KD}, Figure~\ref{fig:weighted_kd}, which scales the logits of the incorrect teachers by hyperparameter W < 1 and distributes the remaining weight to the correct teachers. The final weighting is computed such that to preserve the logits average value range. This way the student can focus on the information from the correct teachers, details in Algorithm~\ref{weighted_kd_alg}.

\begin{algorithm}[H]
\small
\caption{ROaD Weighted KD Normalization}
\begin{algorithmic}
\STATE M : array of models
\STATE input : input features
\STATE target: ground truth targets
\STATE W: scaling hyperparameter
\\\hrulefill
\STATE num\_correct $\gets$ 0
\STATE num\_incorrect $\gets$ 0
\STATE teachers\_logits $\gets$ []
\STATE teachers\_correctness $\gets$ []

\FOR{m in M}
\STATE logits $\gets$ m(input)
\IF {logits = target} 
\STATE num\_correct $\gets$ num\_correct + 1
\STATE teachers\_logits.append(logits)
\STATE teachers\_correctness.append(True)
\ELSE
\STATE num\_incorrect $\gets$ num\_incorrect + 1
\STATE teachers\_logits.append($\textrm{logits} * W$)
\STATE teachers\_correctness.append(False)
\ENDIF
\ENDFOR
 
\STATE $\hat{W} \gets (\textrm{size(M)} - (\textrm{num\_incorrect} * W)) / \textrm{num\_correct}$
 
\FOR{i in 1, 2, size(M)}
\IF {teachers\_correctness[i] = True}
\STATE teachers\_logits[i] $\gets$ teachers\_logits[i] * $\hat{W}$
\ENDIF
\ENDFOR
\STATE final\_teacher\_logits $\gets$ sum(teachers\_logits) / size(M)

\end{algorithmic}
\label{weighted_kd_alg}
\end{algorithm}

\section{Experiments}

\textbf{Evaluation Datasets.} We evaluate on SQUAD 2.0\footnote[1]{We use the public dev set for evaluation}, MNLI matched, QQP, and NewsQA~\cite{trischler2016newsqa}. SQUAD 2.0 is not in our MTL  datasets (Table~\ref{mtl_dataset}) which severs to evaluate generalization for non-pretraining datasets. Moreover, we apply the SQUAD 2.0 trained model to the NewsQA dev set to evaluate robustness to out-of-domain and out-of-distribution data. MNLI and QQP are part of MTL pretraining datasets. All datasets are in English.

\textbf{Implementation.} Our implementation is based on PyTorch. We use the publicly available BERT, ALBERT~\citep{lan2019albert} and ELECTRA models as initial self-supervised pretrained models. Details about the hyperparameters are in the Appendix. We do not perform any preprocessing on the datasets.

\textbf{Metrics.} We use the standard metrics for their respective datasets: F1 score, exact match, and accuracy. To account for model variations we train five times and average the results.

\subsection{Multi-task Pretraining Results}

We compare ROaD MTL pretraining with MT-DNN using our selection of datasets. Table~\ref{mtl_results} shows that ROaD MTL consistently outperforms MT-DNN for ALBERT and ELECTRA. Note that MT-DNN improves only the BERT model, we hypothesis the reason is that BERT is less optimized since it was trained for significantly smaller number of steps and on smaller training data compared to the other models. In addition, we can see in Table~\ref{tricks} that even simple techniques like label soothing~\citep{muller2019does} and data augmentation~ with negative questions improve BERT but not ALBERT and ELECTRA.

Improving upon extremely optimized models, like ELECTRA and ALBERT that already found a good local-minima, is difficult. Moving such a model in the direction of a single task is insufficient to improve the model's overall performance. However, averaging the gradient from different MTL tasks prevents the model from wandering in the wrong direction, i.e., a direction that is beneficial for a single task. Thus, we attribute the ROaD MTL improvements, even with ELECTRA and ALBERT, to the gradient averaging step across tasks.

\begin{table}[!ht]
\centering
\small
\setlength{\tabcolsep}{0.5em} 
{\renewcommand{\arraystretch}{1.3}
\begin{tabular}{lcccc}
\hline
\textbf{Model} & \multicolumn{2}{c}{\textbf{SQUAD}\**} & \textbf{MNLI-m} & \textbf{QQP} \\ \hline
 & \textbf{F1} & \textbf{EM} & \textbf{Acc} & \textbf{Acc} \\ \hline
BERT-large & 84.9 & 81.9 & 86.3 & 91.1 \\
\hspace{1 mm}w/ MT-DNN  & 85.5 & 82.5 & 87.0 & 91.9 \\
\hspace{1 mm}w/ ROaD MTL  & 85.6 & 82.5 & 87.2 & 92.1 \\ \hline
ALBERT-xlarge & 87.9 & 85.0 & 87.9 & - \\
\hspace{1 mm}w/ MT-DNN & 88.0 & 85.0 & 87.9 & - \\
\hspace{1 mm}w/ ROaD MTL  & 88.3 & 85.4 & 88.3 & - \\ \hline
ELECTRA-large & 90.5 & 87.9 & 90.9 & 92.2 \\
\hspace{1 mm}w/ MT-DNN & 90.3 & 87.5 & 90.8 & 92.1 \\
\hspace{1 mm}w/ ROaD MTL & \textbf{90.9} & \textbf{88.3} & \textbf{91.2} & \textbf{92.4} \\
\hline
\end{tabular}
}
\caption{\label{mtl_results}
ROaD MTL vs MT-DNN results
}
\end{table}

\begin{table}[!ht]
\centering
\small
\setlength{\tabcolsep}{0.5em} 
{\renewcommand{\arraystretch}{1.3}
\begin{tabular}{lccc}
\hline
\textbf{Technique} & \textbf{BERT} & \textbf{ALBERT} & \textbf{ELECTRA} \\ \hline
Baseline & 84.9 & 87.9 & 90.5 \\
Label smoothing & 85.2 & 88.2 & 90.4 \\
Data augmentation & 85.6 & 88.2 & 90.4 \\
\hline
\end{tabular}
}
\caption{\label{tricks}
Impact of label soothing and data augmentation on the SQUAD 2.0 dev F1. The augmentation is with negative questions generated by replacing the entity in the question.
}
\vspace{-3mm} 
\end{table}

\subsection{Knowledge Distillation in Multi-task Pretraining Results}

\begin{table}[!ht]
\centering
\small
\setlength{\tabcolsep}{0.5em} 
{\renewcommand{\arraystretch}{1.3}
\begin{tabular}{lcccccc}
\hline
\textbf{Model} & \multicolumn{2}{c}{\textbf{SQUAD}\**} & \multicolumn{2}{c}{\textbf{NewsQA}\**} & \textbf{MNLI-m} &\textbf{QQP} \\ \hline
 & \textbf{F1} & \textbf{EM} & \textbf{F1} & \textbf{EM} & \textbf{Acc} & \textbf{Acc} \\
Baseline & 90.5 & 87.9 & 55.6 & 43.4 & 90.9 & 92.2 \\
+ MTL & 90.9 & 88.3 & 56.8 & 43.7 & 91.2 & 92.4 \\
+ KD & \textbf{91.0} & \textbf{88.5} & \textbf{57.9} & \textbf{46.1} & \textbf{91.4} & \textbf{92.6} \\ \hline
\end{tabular}
}
\caption{\label{model_performance}
ROaD with KD during MTL results. The baseline is ELECTRA-large. The last row is KD during MTL. NewsQA predictions are from the SQUAD model.
}
\vspace{-4mm} 
\end{table}

\textbf{KD during MTL.} Table~\ref{model_performance} shows the results of ROaD MTL pretraining with and without KD. We set the KD hyperparameters temperature $T=5$ and $\lambda = 0.7$. Compared to the baseline ELECTRA-large, the use of ROaD MTL pretraining improves the model performance and robustness across the board. Using KD during the MTL pretraining step with regular finetuning improves the model performance even further since it provides a richer pretraining signal. In particular, on the out-of-domain and out-of-distribution NewsQA dataset. Which demonstrates that models trained with KD during MTL pretraining were able to learn better universal language representation.

In addition, the MNLI-m \textbf{91.4} and QQP \textbf{92.6} results are new best published accuracies for a single model. 

\subsection{Knowledge Distillation in Finetuning Results}
To apply KD during task finetuning, we trained three ROaD MTL models on SQUAD 2.0 using different random seeds. Then, we used them as a teacher ensemble for the final student model by collecting and averaging their logits as soft targets. We searched for the best distillation hyperparameters and found a temperature $T=5$ and $\lambda = 0.9$ worked best.

We compare Weighted KD to KD and the recent teacher annealing method. Teacher annealing~\citep{clark2019bam} mixes the teacher soft targets with the ground truth hard targets, where the weight of the hard targets is linearly increased from 0 to 1 throughout training. Early in training, the model is distilling from the teacher. Towards the end of training, the model relies on the hard targets.

Table~\ref{comparing_kd} shows that KD improves upon a single model and almost matches the ensemble while teacher annealing is lower. Weighted KD outperforms the traditional logits KD and teacher annealing, and the student surpassed the teacher ensemble performance. This shows that Weighted KD can leverage the strengths of the individual teachers.


\begin{table}[!ht]
\centering
\small
\setlength{\tabcolsep}{0.5em} 
{\renewcommand{\arraystretch}{1.3}
\begin{tabular}{lcc}
\hline
 & \textbf{F1} & \textbf{EM} \\ \hline
Single model average w/o distillation & 90.86 & 88.35 \\
Teachers ensemble \textbackslash{}w 3 models & 91.41 & 88.90 \\
KD & 91.38 & 88.84 \\
Annealing KD & 91.0 & 88.1 \\
\textbf{Weighted KD (this work)} & \textbf{91.52} & \textbf{88.96} \\ \hline
\end{tabular}
}
\caption{\label{comparing_kd}
KD during finetuning with ELECTRA-large on SQUAD 2.0 dev.
}
\vspace{-4mm} 
\end{table}

\subsection{ROaD Results}

We use the ROaD modeling framework to train an ELECTRA-large MRC model -- KD during MTL pretraining and weighted KD in finetuning,. We compare with the Retro-Reader~\citep{zhang2020retrospective} MRC model which at this time is the best published result. Retro-Reader integrates two ELECTRA models into two stages of reading and verification (1) sketchy reading that briefly investigates the overall interactions between passage and question (2) intensive reading that verifies the answer and gives the final prediction. 

Table~\ref{final_result} and Table~\ref{final_result_base} shows the ROaD large and ROaD base results together with leading single models on SQUAD 2.0 dev and out-of-domain and out-of-distribution on NewsQA dev. We we make the following observations:
\begin{itemize}
\item ROaD outperforms strong models such as ELECTRA, DeBERTa~\citep{he2020deberta} and Retro-Reader, despite the Retro-Reader contains two ELECTRA-large models, sketchy and intensive. This is a new state-of-the-art for SQUAD 2.0 devset.
\item ROaD model surpassed the performance of ensemble of three ELECTRA-large models.
\item ROaD model improves robustness and generalizes better on the out-of-domain NewsQA dataset. 
\end{itemize}

\begin{table}[!ht]
\centering
\small
\setlength{\tabcolsep}{0.5em} 
{\renewcommand{\arraystretch}{1.3}
\begin{tabular}{lllll}
\hline
 & \multicolumn{2}{l}{\textbf{SQUAD}\**} & \multicolumn{2}{l}{\textbf{NewsQA}\**} \\ \hline
 & \textbf{F1} & \textbf{EM} & \textbf{F1} & \textbf{EM} \\ \hline
BERT-large & 87.5 & 84.7 & - & - \\
RoBERTa-large & 89.4 & 86.5 & - & - \\
DeBERTa-large & 90.7 & 88.0 & - & - \\
ELECTRA-large & 90.5 & 87.9 & 55.6 & 43.4 \\
\hspace{1 mm}w/ 3 models ensemble & 90.8 & 88.3 & 56.0 & 43.7 \\
\hspace{1 mm}w/ Retro-Reader & \underline{91.3} & \underline{88.8} & - & - \\
\hspace{1 mm}w/ ROaD & \textbf{91.6} & \textbf{89.1} & \textbf{58.0} & \textbf{46.2} \\ \hline
\end{tabular}
}
\caption{\label{final_result}
Large model results for SQUAD 2.0 and NewsQA. Bold is the new state-of-the-art and underline is the previous best result.
}
\vspace{-4mm} 
\end{table}

\begin{table}[!ht]
\centering
\small
\setlength{\tabcolsep}{0.5em} 
{\renewcommand{\arraystretch}{1.3}
\begin{tabular}{lllll}
\hline
 & \multicolumn{2}{l}{\textbf{SQUAD}\**}  \\ \hline
 & \textbf{F1} & \textbf{EM} \\ \hline
RoBERTa-base & 83.7 & 80.5 \\
DeBERTa-base & \underline{86.2} & \underline{83.1} \\
ELECTRA-base & 83.3 & 80.5 \\
\hspace{1 mm}w/ ROaD & \textbf{87.6} & \textbf{85.1} \\ \hline
\end{tabular}
}
\caption{\label{final_result_base}
Base model results for SQUAD 2.0. Bold is the new state-of-the-art and under-line is the previous best result.
}
\vspace{-4mm} 
\end{table}
\section{Conclusion}
We introduced the ROaD modeling framework which combines KD with MTL during pretraining and KD from multiple teachers during finetuning. Experiments show that ROaD-ELECTRA model for MRC and NLI yields significant improvements over strong baselines and achieves new state-of-the-art for a single model.

In future work we would like to explore weighted KD with ensemble of diverse models such as ELECTRA, ALBERTA, and BERT.

\section{Supplementary Material}

\subsection{Hardware}

We use 8 NVIDIA V100 GPUs with 32GB memory. We train MTL for total of 50,000 steps which takes about 12 hours on 8 GPUs.

\subsection{Hyperparamters}

We use the hyperparamters in Table~\ref{trainign_hyper} for pretraining and finetunning the models.

\begin{table}[!ht]
\centering
\small
\setlength{\tabcolsep}{0.15em} 
{\renewcommand{\arraystretch}{1.2}
\begin{tabular}{lll}
\hline
\textbf{Hyperparameter} & \textbf{Pretraining} & \textbf{Finetuning} \\ \hline
Steps & 50,000 & - \\
Epochs & - & 3 \\
Warmup steps & 10\% & 10\% \\
Learning rate & 3e-4 & 5e-5 \\
Layer-wise lr multiplier & 0.75 & 0.9 \\
Batch size & 512 & 32 \\
Training examples per pass & 16 & - \\
Adam $\epsilon$ & 1e-6 & 1e-6 \\
Adam $\beta_1$ & 0.9 & 0.9 \\
Adam $\beta_1$ & 0.999 & 0.999 \\
Attention dropout & 0.1 & 0.1 \\
Dropout & 0.1 & 0.1 \\
Weight decay & Linear & Linear \\
Gradient clipping & 1.0 & 1.0
\end{tabular}
}
\caption{\label{trainign_hyper}
Training hyperparamters
}
\end{table}

For KD, we tried different values for $T = \{7, 5, 3\}$ and $\lambda = \{0.5, 0.7, 0.9\}$. We did not notice significant differences but in our experiments $T = 5$ and $\lambda = 0.9$ had the best results. Similarity, for weighted KD, we tried $W = {0.7, 0.75, 0.8}$ and $W = 0.75$ worked best but with not significant differences.

Table~\ref{model_sizes} shows the size of the model we used.

\begin{table}[!ht]
\centering
\small
\setlength{\tabcolsep}{0.5em} 
{\renewcommand{\arraystretch}{1.2}
\begin{tabular}{ll}
\hline
\textbf{Model} & \textbf{\# Parameters} \\ \hline
BERT-large & 335M \\
RoBERTa-large & 355M \\
ELECTRA-large & 355M \\
DeBERTa-large & 400M \\
ALBERT-xlarge & 58M \\
\end{tabular}
}
\caption{\label{model_sizes}
Model sizes
}
\end{table}

\bibliography{anthology}
\bibliographystyle{acl_natbib}

\end{document}